\pdfoutput=1
\documentclass[10pt, a4paper]{article}
\usepackage{lrec2022} 
\usepackage{multibib}
\newcites{languageresource}{Language Resources}
\usepackage{graphicx}
\usepackage{tabularx}
\usepackage{soul}
\usepackage{placeins}

\usepackage{titlesec}
\titleformat{\section}{\normalfont\large\bfseries\center}{\thesection.}{1em}{}
\titleformat{\subsection}{\normalfont\SmallTitleFont\bfseries\raggedright}{\thesubsection.}{1em}{}
\titleformat{\subsubsection}{\normalfont\normalsize\bfseries\raggedright}{\thesubsubsection.}{1em}{}
\renewcommand\thesection{\arabic{section}}
\renewcommand\thesubsection{\thesection.\arabic{subsection}}
\renewcommand\thesubsubsection{\thesubsection.\arabic{subsubsection}}

\usepackage{epstopdf}
\usepackage[utf8]{inputenc}

\usepackage{hyperref}
\usepackage{xstring}

\usepackage{color}

\usepackage{microtype}
\usepackage{booktabs}
\usepackage{makecell}

\usepackage[para]{footmisc}

\usepackage{bera}
\usepackage{listings}
\usepackage{xcolor}

\definecolor{eclipseStrings}{RGB}{42,0.0,255}
\definecolor{eclipseKeywords}{RGB}{127,0,85}
\colorlet{numb}{magenta!60!black}

\lstdefinelanguage{python}{
 keywords={typeof, null, catch, switch, in, int, str, float, self},
 keywordstyle=\bfseries,
 ndkeywords={boolean, throw, import, return, class, if, elif, endif, while, do, else, True, False , catch, def},
 ndkeywordstyle=\bfseries,
 identifierstyle=\color{black},
 sensitive=false,
 comment=[l]{\#},
 morecomment=[s]{/*}{*/},
 commentstyle=\color{purple}\ttfamily,
 string=[s]{"}{"},
 stringstyle=\ttfamily,
}

\colorlet{punct}{red!60!black}
\definecolor{background}{HTML}{EEEEEE}
\definecolor{delim}{RGB}{20,105,176}
\colorlet{numb}{magenta!60!black}

\lstdefinelanguage{json}{
    basicstyle=\normalfont\ttfamily,
    numbers=left,
    numberstyle=\scriptsize,
    stepnumber=1,
    numbersep=8pt,
    showstringspaces=false,
    breaklines=true,
    frame=lines,
    literate=
     *{0}{{{\color{numb}0}}}{1}
      {1}{{{\color{numb}1}}}{1}
      {2}{{{\color{numb}2}}}{1}
      {3}{{{\color{numb}3}}}{1}
      {4}{{{\color{numb}4}}}{1}
      {5}{{{\color{numb}5}}}{1}
      {6}{{{\color{numb}6}}}{1}
      {7}{{{\color{numb}7}}}{1}
      {8}{{{\color{numb}8}}}{1}
      {9}{{{\color{numb}9}}}{1}
      {:}{{{\color{punct}{:}}}}{1}
      {,}{{{\color{punct}{,}}}}{1}
      {\{}{{{\color{delim}{\{}}}}{1}
      {\}}{{{\color{delim}{\}}}}}{1}
      {[}{{{\color{delim}{[}}}}{1}
      {]}{{{\color{delim}{]}}}}{1},
}

\lstdefinelanguage{jsonblue}{
    basicstyle=\scriptsize,
    commentstyle=\color{black}, 
    stringstyle=\color{blue}, 
    showstringspaces=false,
    breaklines=true,
    breakatwhitespace=true,
    frame=lines,
    string=[s]{"}{"},
    comment=[s]{:\ "}{"},
    morecomment=[s]{:\ ["}{]},
    literate=
        *{0}{{{\color{numb}0}}}{1}
         {1}{{{\color{numb}1}}}{1}
         {2}{{{\color{numb}2}}}{1}
         {3}{{{\color{numb}3}}}{1}
         {4}{{{\color{numb}4}}}{1}
         {5}{{{\color{numb}5}}}{1}
         {6}{{{\color{numb}6}}}{1}
         {7}{{{\color{numb}7}}}{1}
         {8}{{{\color{numb}8}}}{1}
         {9}{{{\color{numb}9}}}{1}
         {type}{{{\color{blue}type}}}{1}
}
\lstdefinelanguage{jsongreen}{
    basicstyle=\scriptsize,
    commentstyle=\color{black}, 
    stringstyle=\color{green}, 
}

\newenvironment{citemize}{\begin{list}{$\bullet$}{\topsep=.2\smallskipamount\itemsep=0pt\parsep=1pt\labelwidth=.5em}}{\end{list}}
\newenvironment{cenumerate}{\begin{list}{\labelenumi}{\usecounter{enumi}\topsep=.2\smallskipamount\itemsep=0pt\parsep=1pt\labelwidth=.5em}}{\end{list}}

\title{DaMuEL: A Large Multilingual Dataset for Entity Linking}

\name{David Kubeša, Milan Straka} 

\address{Charles University, Faculty of Mathematics and Physics\\
         Institute of Formal and Applied Linguistics\\
         \{kubesa, straka\}@ufal.mff.cuni.cz\\}

\abstract{We present DaMuEL, a large \textbf{Mu}ltilingual \textbf{Da}taset for
\textbf{E}ntity \textbf{L}inking containing data in 53 languages. DaMuEL consists of two components: a knowledge base that contains language-agnostic information about entities, including their claims from Wikidata and named entity types (PER, ORG, LOC, EVENT, BRAND, WORK\_OF\_ART, MANUFACTURED); and Wikipedia texts with entity mentions linked to the knowledge base, along with language-specific text from Wikidata such as labels, aliases, and descriptions, stored separately for each language. The Wikidata QID is used as a persistent, language-agnostic identifier, enabling the combination of the knowledge base with language-specific texts and information for each entity.
Wikipedia documents deliberately annotate only a single
mention for every entity present; we further automatically detect all mentions
of named entities linked from each document. The dataset contains 27.9M named entities in the knowledge base and 12.3G tokens from Wikipedia texts. The dataset is published under the CC BY-SA license at {\small\url{https://hdl.handle.net/11234/1-5047}}.
 \\ \newline  }

\begin{document}

\maketitleabstract

\section{Introduction}

Recognition of named entity mentions is an important step in various
NLP applications. Both named entity recognition (NER), which annotates spans
and types of named entity mentions in a given text, and more complex
named entity linking (NEL), which links each named entity mention to
a provided knowledge base, are beneficial for example for
dialogue systems (``who was the father of Charles IV''), information retrieval (``find documents about
Charles IV''), or question answering, to name just a few.

Nowadays, several pipelines like UDPipe~\cite{straka-2018-udpipe} or
Stanza~\cite{qi-etal-2020-stanza} offers multilingual morphosyntactic analysis
for nearly a hundred languages, based on the Universal Dependencies
project \cite{ud}. For NER and NEL, however, the situation is different.
Even if these tasks are extensively studied, not many datasets are
available in other than the English language.

In this paper, we present DaMuEL, a large \textbf{Mu}ltilingual \textbf{Da}taset for
\textbf{E}ntity \textbf{L}inking. It provides data in 53 languages extracted from
Wikidata\footnote{\scriptsize\url{https://www.wikidata.org}} and
Wikipedia\footnote{\scriptsize\url{https://www.wikipedia.org}}. 
DaMuEL consists of the following components:

\begin{citemize}
\item A \textit{knowledge base} containing language-agnostic information about entities, including their claims from Wikidata and named entity types (PER, ORG, LOC, EVENT, BRAND, WORK\_OF\_ART, MANUFACTURED)\rlap{.}
    \item \textit{Texts} from Wikipedia with entity mentions linked to the knowledge base, along with language-specific text from Wikidata such as labels, aliases, and descriptions, stored separately for each language.
\end{citemize}

The dataset has the following distinctive properties:
\begin{citemize}
  \item It is a massive multilingual dataset, containing on average more
    than 233.0M words and 507.3k named entities that have Wiki page in each of the 53 languages.
  \item The entities utilize the Wikidata Q identifier (QID) as a persistent
    language-agnostic identifier, which allows a combination of the knowledge base with language-specific texts and information for each entity.
  \item For named entities, the knowledge base contains an automatically
    determined type (PER, ORG, LOC, EVENT, BRAND, WORK\_OF\_ART, MANUFACTURED), according to the hierarchy of entities in
    Wikidata.
  \item The Wikipedia texts are tokenized and morphologically analyzed using
    UDPipe~\cite{straka-2018-udpipe}.
  \item Because the Wikipedia pages deliberately annotate only a single
    mention for every entity present, we further automatically detect all
    mentions of named entities. We are adding only mentions of entities that are already annotated on each page.
\end{citemize}

The dataset is published under the CC BY-SA licence at {\small\url{https://hdl.handle.net/11234/1-5047}}.

\section{Related Work}

\begin{figure*}
    \centering
    \includegraphics[width=.81\hsize]{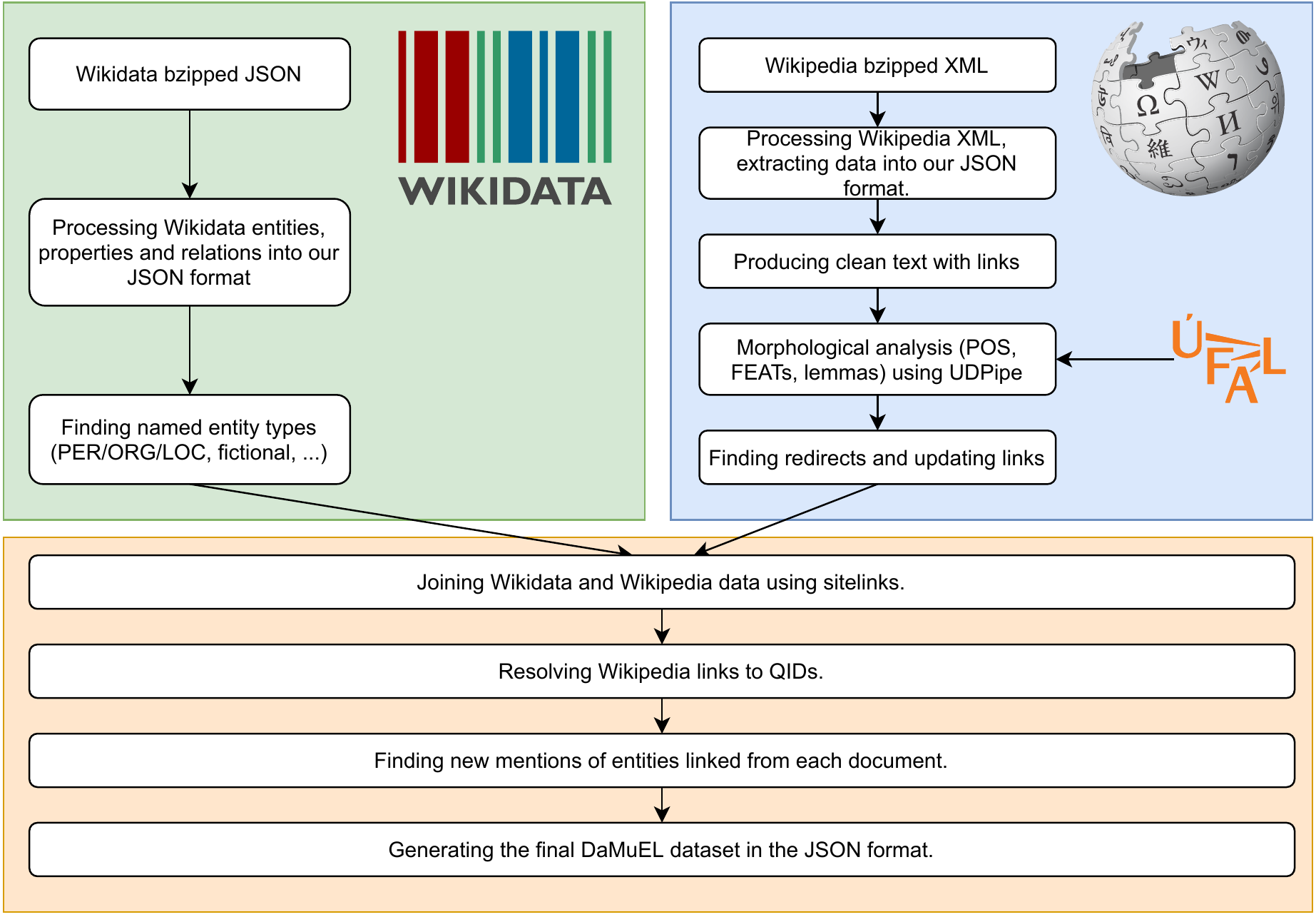}
    \caption{Illustration of the DaMuEL generation process.}
    \label{fig:processing_graph}
\end{figure*}

Given that both named entity recognition and linking are extensively studied, several named entity linking dataset exists, most often for English.

    \textbf{AIDA CoNLL-YAGO Dataset} \cite{hoffart-etal-2011-robust} is based on the frequently-used English NER dataset from CoNLL~\cite{conll2003}. Every original annotated named entity mention is linked to an entity from a knowledge base, using YAGO2 entity name, a Wikipedia URL, and Freebase mid as entity identification.

    \textbf{AIDA-EE Dataset} \cite{hoffart2014discovering} annotates the so-called \textit{emergent entities}, which are new entities not present in a knowledge base. Concretely, 300 documents are annotated with nearly 10k entity mentions, with 561 emergent entities (not present in the Wikipedia Aug 17, 2010 dump).

    The Text Analysis Conference (TAC) is a workshop aiming at improving various NLP applications. It consists of a number of tracks, one of them being Knowledge Base Population, which is organized since 2010~\cite{ji2010overview}. This track includes an entity linking task, which has been used widely for entity linking systems evaluation and for which a number of resources has been published, even in other than English languages:
    \begin{citemize}
        \item \textbf{TAC KBP Knowledge base}~\cite{tackbpKB} containing 800k Wikipedia entities,
        \item \textbf{TAC KBP English Entity Linking}~\cite{tackbpEEL},
        \item \textbf{TAC KBP Spanish Cross-lingual Entity Linking}~\cite{tackbpSEL},
        \item \textbf{TAC KBP Chinese Cross-lingual Entity Linking}~\cite{tackbpCEL}.
    \end{citemize}
    


    \textbf{EDRAK: Entity-Centric Data Resource for Arabic Knowledge} \cite{gad2015edrak} is another example of non-English dataset, consisting of more than 2M entities with Arabic named and contextual keyphrases, generated from Wikipedia.

\section{Dataset Creation}

We now describe the process of creating DaMuEL, which is illustrated in Figure~\ref{fig:processing_graph}. Given that both the resulting dataset and the input sources are huge,
we employ distributed processing via Spark~\cite{spark} during dataset generation.

\subsection{Wikidata}

Wikidata is a free and open knowledge base that acts as structured-data central storage for Wikipedia and related projects. Generally, it is a collection of items, each identified by a persistent \textit{Q identifier}, or \textit{QID} for short.

\subsubsection{Wikidata Structure}
In addition to \textit{QID}, each item can have claims and language-specific properties: label, aliases, and description.

Claims in Wikidata represent a statement about an entity, such as its date of birth, occupation, or location. Each claim consists of a property, a value, and an optional qualifier. For example, the claim ``Albert Einstein was born on March 14, 1879'' consists of the property ``date of birth'', the value ``March 14, 1879'', and the entity ``Albert Einstein''.

Statements in Wikidata are the individual pieces of data that make up a claim. They consist of a value, a property, and an optional qualifier. For example, the statement ``March 14, 1879'' is the value of the property ``date of birth'' for the entity ``Albert Einstein''.

Wikidata also allows for qualifiers to be added to claims. Qualifiers provide additional context or details about the claim, such as the specific time period or location. For example, a qualifier might be added to the claim ``Albert Einstein lived in Princeton'' to indicate the specific years he lived there.

Ranks in Wikidata are used to indicate the degree of confidence in a particular statement. There are three possible ranks: preferred, normal, and deprecated. A preferred rank indicates the highest level of confidence in a statement, while a deprecated rank indicates that the statement is no longer considered accurate or relevant. Normal rank is used for all other statements.

References in Wikidata are used to provide evidence or sources for the claims made about an entity. A reference consists of a source and a set of properties that describe how the source supports the claim. For example, a reference for the claim ``Albert Einstein was born on March 14, 1879'' might include a link to a birth certificate as the source and properties such as the date the source was accessed and the name of the person who added the reference.

\subsubsection{DaMuEL KB Structure}

Our language-agnostic knowledge base includes information from all claims and ranks present in Wikidata. We do not include references in our dataset, as these are not useful for our entity-linking task. Additionally, we annotate each entity in our dataset with information about whether it is a named entity and, if so, which entity type it belongs to. A detailed JSON schema that defines the exact structure of our dataset can be found in Figure\ref{fig.wikidata_structure}.

In the shared knowledge base, there are no language-specific texts, therefore, a label, several aliases, and a description are stored together with the Wikidata texts separately per language for our 53 supported languages. More about that is in Section \ref{section:wikidata}.

\subsubsection{Determining Named Entity Types for the KB}
We base our identification of named entities on the Wikidata claims ``Instance of'', ``Subclass of'', and ``Fictional analog of''.

\begin{citemize}
    \item ``Instance of'': This claim describes the type or class to which an entity belongs. For example, the entity ``Albert Einstein'' has the ``instance of'' claim ``human''. This indicates that Albert Einstein is an instance of the class ``human''.
    \item ``Subclass of'': This claim describes the hierarchical relationship between classes or types. For example, the class ``archaeologist'' has the ``subclass of'' claim ``cultural studies scholar''. This indicates that ``archaeologist'' is a subclass of the more general class ``cultural studies scholar''.
    \item ``Fictional analog of'': This claim is used to indicate that an entity has a fictional counterpart or equivalent. For example, the entity ``Sherlock Holmes'' might have the ``fictional analog of'' claim ``consulting detective''. This indicates that Sherlock Holmes is a fictional character who is a consulting detective.
\end{citemize}
 These three claims are important for organizing and linking information in Wikidata. By using the ``instance of'' and ``subclass of'' claims, it is possible to create a taxonomy of classes and types that can be used to classify and categorize entities.

In order to identify named entities of a particular type, we employ a hierarchical approach that leverages the structure of the Wikidata knowledge graph. Specifically, we look for entities that can be represented by one instance of a specific class and one or more subclasses of the general concept associated with that entity type. For example, to identify a named entity of the PERSON (PER) type, we look for entities that are instances of the Q5 (human) class or any of its subclasses in the Wikidata hierarchy.

In addition, we are interested in identifying named entities that exist in fictional or imaginary contexts, such as characters from literature, film, or video games. We determine such entities by considering entities that are instances of classes that are a ``Fictional analog of'' any subclasses in the Wikidata hierarchy for a given named entity type.

However, there is one exception to this approach. The named entity type MANUFACTURED is associated with entities that have a specific property claim, P176 (manufacturer), rather than a specific class or subclass in the Wikidata hierarchy. This is because the MANUFACTURED entity type is focused on manufactured products, rather than entities that are inherently part of a broader category, such as humans or organizations.

The exhaustive list of the named entity types we recognize, together with the corresponding root entity (or claim) used to identify them, follows:
\begin{citemize}
    \item PER - a person. Wikidata general concept: Q5 (human);
    \item ORG - an organization. Wikidata general concept: Q43229 (organization);
    \item LOC - a geographic location, including a geographic entity. Wikidata general concept: Q27096213 (geographic entity);
    \item EVENT - temporary and scheduled happening, like a conference, festival, competition, or similar.Wikidata general concept: Q1656682 (event);
    \item BRAND - identification for a good or service. Wikidata general concept: Q431289 (brand);
    \item WORK\_OF\_ART - an aesthetic item or artistic creation. Wikidata general concept: Q838948 (work of art);
    \item MANUFACTURED - some manufacturer or producer manufactures it. Wikidata property claim: P176 (manufacturer).
\end{citemize}

\subsection{Wikipedia}
\label{section:wikidata}

Wikipedia is a collection of documents edited by humans. The documents are represented using a specific markup language called \textit{wikitext}. The Wikipedia markup is quite difficult to process for several reasons:
\begin{citemize}
\item there is no formal standardization;
\item the markup can be language-specific;
\item because Wikipedia is edited by humans, there are various errors in the markup.
\end{citemize}
Furthermore, wikitext extensively utilizes \textit{templates}, which can be considered macros, possibly with one or more inputs.

To process the wikitext, we use an existing Python parser \textit{MediaWiki Parser from Hell}.\footnote{\scriptsize\url{https://github.com/earwig/mwparserfromhell}} Our goal is to provide manually created plain texts with links to other entities. We therefore:
\begin{citemize}
  \item ignore the templates;
  \item remove lists, tables, images, and galleries;
  \item remove links to categories;
  \item remove in-line references;
  \item remove in-line formatting markup.
\end{citemize}
We do not currently include templates. However, templates may contain valuable information, such as units of measurement. As part of our future work, we plan to improve the handling of templates. It should be noted that expanding templates alone will not be sufficient, and additional processing will be necessary.
Note that we keep the information that some textual content (like lists or tables) was deleted.

After processing, each document is a collection of plain text paragraphs and entities linked from their mentions. We then tokenize the paragraphs and perform morphological analysis using UDPipe~\cite{straka-2018-udpipe}, representing each paragraph as a collection of sentences and each sentence as a collection of tokens with lemmas, UPOS (universal part-of-speech tags), and FEATS (detailed morphological features)~\cite{ud}.

Finally, to prepare for link resolution, we need to process redirects -- a Wikipedia page can act only as a (potentially recurrent) redirect to another document. We, therefore, collect the redirect information and replace links to redirects with their targets.

\subsection{Joining Wikidata and Wikipedia}

We now join the preprocessed Wikidata and Wikipedia. Conveniently, Wikidata items contain \textit{sitelinks}, which are pointers to language-specific Wikipedia pages about the corresponding Wikidata item. This provides the QID identifiers for the Wikipedia documents and allows us to resolve Wikipedia links (references to other Wikipedia pages) into QIDs.

\subsection{Expanding Entity Mentions}
\label{section.adding_mentions}

In Wikipedia, an entity should be generally linked only once from a given document. To quote from the Wikipedia Manual of Style:\footnote{\scriptsize\url{https://en.wikipedia.org/wiki/Wikipedia:Manual_of_Style/Linking}}

{
\setlength{\leftskip}{1em}
\textit{Generally, a link should appear only once in an article, but it may be repeated if helpful for readers, such as in infoboxes, tables, image captions, footnotes, hatnotes, and at the first occurrence after the lead.}

}

Nevertheless, in order to use DaMuEL for training a NER/NEL systems, we would like all entity mentions to be annotated. Therefore, we \textit{expand} the mentions in every document by locating other mentions of already linked named entities (we chose only to consider the already annotated named entities to avoid many false positives). To match other mentions of the linked entities, we consider the following candidates, with priority from top to bottom:

\begin{cenumerate}
\label{priority}
    \item Wikidata labels,
    \item Wikidata aliases,
    \item Wikipedia titles,
    \item Wikipedia redirect texts,
    \item Wikipedia anchor texts (i.e., the entity mention texts).
\end{cenumerate}

We match the candidates and texts by an exact match and a lemma match. For matching, we remove the parentheses from the titles used for disambiguation. The anchor texts are stripped of commas and special characters.

We record all discovered candidates in the dataset, including their origin. However, some of the candidates can overlap. Therefore, we additionally employ a heuristic, whose goal is to choose a non-overlapping \textit{flat} set of candidates. The heuristic prefers longer mentions, and if several mentions are of the same length, the one with the highest priority (according to the above list) is selected. For every token, we select its best match independently, and a mention is chosen to the \textit{flat} set only if it has been selected by its every word.

The anchors in Wikipedia proved to be quite noisy. We added these conditions to the anchor texts to be considered:
\begin{citemize}
    \item There must be at least 10 occurrences of the given anchor text in Wikipedia to include this anchor for our search.
    \item Only mentions that represent over 10 percent of occurrences of the anchor texts of that entity are used for the expansion. This means that every entity can have maximally 10 variants of anchor texts.
\end{citemize}

\section{The DaMuEL Dataset}

\begin{table}[t]
    \centering
    \begin{tabular}{lrr}
    \toprule
      & \makecell{Average\\per Language} & Total~~ \\
    \midrule
        Pages & 725.5k & ~~~--- \\
        Tokens & 233.0M & 12.3G \\
        Sentences & 12.0M & 637.2M \\
        Paragraphs & 5.3M & 278.9M \\
        \makecell[l]{Named entities\\with wiki page} & 507.3k & ~~~--- \\
        Entities & 4.7M & ~~~--- \\
        Named entities & 3.4M & ~~~--- \\
        \makecell[l]{Disk size\\per language} & 26.1GB & 1.4TB \\
        Wikidata size & 12.8GB & ~~~--- \\
    \bottomrule
    \end{tabular}
    \caption{DaMuEL overall statistics across all 53 languages.}
    \label{tab:statistics}
\end{table}

\begin{figure*}
\raggedleft
\begin{lstlisting}[language=jsonblue,basicstyle=\linespread{0.97}\scriptsize]
{
  "$schema": "https://json-schema.org/draft/2020-12/schema",
  "description": "DaMuEL dataset - KB",
  "type": "object",
  "$defs": {
    "typed_value": {
      "oneOf": [
        {"prefixItems": [{"const": "qid"}, {"pattern": "^Q\\d+($|:)"}]},
        {"prefixItems": [{"const": "url"}, {"type": "string"}]},
        {"prefixItems": [{"const": "string"}, {"type": "string"}]},
        {"prefixItems": [{"const": "external-id"}, {"type": "string"}]},
        {"prefixItems": [{"const": "wikibase-item"}, {"type": "string"}]},
        {"prefixItems": [{"const": "wikibase-property"}, {"type": "string"}]},
        {"prefixItems": [{"const": "commonsMedia"}, {"type": "string"}]},
        {"prefixItems": [{"const": "geo-shape"}, {"type": "string"}]},
        {"prefixItems": [{"const": "math"}, {"type": "string"}]},
        {"prefixItems": [{"const": "musical-notation"}, {"type": "string"}]},
        {"prefixItems": [{"const": "tabular-data"}, {"type": "string"}]},
        {"prefixItems": [{"const": "wikibase-lexeme"}, {"type": "string"}]},
        {"prefixItems": [{"const": "wikibase-form"}, {"type": "string"}]},
        {"prefixItems": [{"const": "wikibase-sense"}, {"type": "string"}]},
        {"prefixItems": [{"const": "globe-coordinate"}, {"type": "string"}]},
        {"prefixItems": [{"pattern": "^monolingualtext:[-a-z]+($|\\d*)"}, {"type": "string"}]},
        {"prefixItems": [{"pattern": "^time:(gregorian|julian)$"}, {"type": "string"}]},
        {"prefixItems": [{"pattern": "^quantity:Q\\d+($|:)"}, {"type": "string"}]}
      ]
    }
  },
  "properties": {
    "qid": {"description": "The unique identifier for an article", "type": "string", "pattern": "^Q\\d+($|:)"},
    "claims": {
      "description": "The set of claims for an article",
      "type": "object",
      "additionalProperties": {
        "type": "array",
        "items": {
          "type": "array",
          "minItems": 2,
          "maxItems": 3,
          "allOf": [
            {"$ref": "#/$defs/typed_value"},
            {
              "prefixItems": [true,true,
                {
                  "type": "object",
                  "additionalProperties": {
                    "type": "array",
                    "items": {"type": "array", "minItems": 2, "maxItems": 2, "$ref": "#/$defs/typed_value"}
    }}]}]}}},
    "named_entities": {
      "description": "The set of named entities for an article",
      "type": "object",
      "properties": {
        "type": {
          "type": "array",
          "items": {"enum": ["PER", "ORG", "LOC", "EVENT", "BRAND", "WORK_OF_ART", "MANUFACTURED"]}
        }
      },
      "fictional": {"description": "If the entity is fictional", "type": "boolean"},
      "required": ["type"]
    }
  },
  "required": ["qid"]
}
\end{lstlisting}
    \caption{DaMuEL JSON structure of the common information from Wikidata. It contain information if the entity is a named entity and all claims coming from Wikidata.}
    \label{fig.wikidata_structure}
\end{figure*}

\begin{figure*}
\raggedleft
\begin{lstlisting}[language=jsonblue,basicstyle=\linespread{0.645}\scriptsize]
{
  "$schema": "https://json-schema.org/draft/2020-12/schema",
  "description": "DaMuEL dataset - texts",
  "type": "object",
  "properties": {
    "qid": {"description": "The unique identifier for an article", "type": "string", "pattern": "^Q\\d+($|:)"},
    "lang": {"description": "The language of the article", "type": "string"},
    "aliases": {"description": "The aliases of the article", "type": "array", "items": {"type": "string"}},
    "description": {"description": "The description of the article", "type": "string"},
    "label": {"description": "The label of the article", "type": "string"},
    "wiki": {
      "type": "object",
      "properties": {
        "title": {"description": "The title of the article", "type": "string"},
        "text": {"description": "The text of the article", "type": "string"},
        "tokens": {
          "description": "The tokens of the article",
          "type": "array",
          "items": {
            "type": "object",
            "properties": {
              "start": {"description": "The start position of the token in the text property", "type": "integer"},
              "end": {"description": "The end position of the token in the text property", "type": "integer"},
              "lemma": {"description": "The lemma of the token", "type": "string"},
              "upostag": {"description": "The part-of-speech of the token", "type": "string"},
              "feats": {"description": "The features of the token", "type": "string"}
            },
            "required": ["start", "end", "lemma", "upostag", "feats"]
          }
        },
        "sentences": {
          "description": "The sentences of the article",
          "type": "array",
          "items": {
            "type": "object",
            "properties": {
              "start": {"description": "The start position of the sentence in tokens", "type": "integer"},
              "end": {"description": "The end position of the sentence  in tokens", "type": "integer"}
            },
            "required": ["start", "end"]
          }
        },
        "paragraphs": {
          "description": "The paragraphs of the article",
          "type": "array",
          "items": {
            "type": "object",
            "properties": {
              "start": {"description": "The start position of the paragraph in tokens", "type": "integer"},
              "end": {"description": "The end position of the paragraph in tokens", "type": "integer"},
              "heading": {"description": "The heading level of the paragraph", "type": "integer"}
            },
            "required": ["start", "end"]
          }
        },
        "redirects": {"description": "The redirects to the article", "type": "array", "items": {"type": "string"}},
        "anchors": {
          "description": "The anchors of the article",
          "type": "object", "additionalProperties": {"type": "integer"}
        },
        "removed_text_before": {
          "description": "The removed text before the token",
          "type": "array", "items": {"type": "integer"}
        },
        "links": {
          "description": "The links of the article",
          "type": "array",
          "items": {
            "type": "object",
            "properties": {
              "start": {"description": "The start position of the link in tokens", "type": "integer"},
              "end": {"description": "The end position of the link in tokens", "type": "integer"},
              "qid": {"description": "The target qid of the link", "type": "string", "pattern": "^Q\\d+($|:)"},
              "origin": {
                "description": "Says if the links was originally in wikipedia or added by us",
                "type": "string",
                "enum": ["wiki", "title", "title_lemmatized", "label", "label_lemmatized", "alias",
                  "alias_lemmatized", "redirects", "redirects_lemmatized", "anchors", "anchors_lemmatized"]
              },
              "title": {"description": "The target title of the link", "type": "string"},
              "flat": {"description": "No word can be in more links, so this link is a dominant one", "type": "boolean"}
            },
            "required": ["start", "end", "title", "origin"]
          }, "additionalProperties": false
        },
        "removed_links": {
          "description": "The removed links of the article",
          "type": "array",
          "items": {
            "type": "object",
            "properties": {
              "count": {"description": "Number of times the link was removed", "type": "integer"},
              "qid": {"description": "The target qid of the link", "type": "string", "pattern": "^Q\\d+($|:)"},
              "title": {"description": "The target title of the link", "type": "string"}
            },
            "required": ["count", "title"]
          }
        }
      },
      "required": ["title", "text", "tokens", "sentences", "paragraphs", "links"], "additionalProperties": false
    }
  },
  "required": ["qid", "lang"], "additionalProperties": false
}
\end{lstlisting}
    \caption{DaMuEL language-specific JSON structure. The descriptions, labels, and aliases come from Wikidata. The data coming from Wikipedia are under the wiki key.}
    \label{fig.wiki_structure}
\end{figure*}

\begin{table*}[!t]
    \centering
    {
    \renewcommand{\arraystretch}{0.94}
    \setlength{\tabcolsep}{10pt}
    \small
    \begin{tabular}{lrrrrr}
    \toprule
      Language & Tokens & Pages & \makecell{Named\\Entities\\with Page} & \makecell{Named\\Entities} & Entities  \\
    \midrule
Afrikaans & 27.4M & 103.4k & 65.3k & 441.2k & 702.8k \\
Arabic & 212.4M & 1143.3k & 893.8k & 9555.4k & 11766.4k \\
Armenian & 68.0M & 273.3k & 214.3k & 2356.3k & 4015.9k \\
Basque & 55.4M & 388.4k & 253.3k & 2462.2k & 4646.5k \\
Belarusian & 39.4M & 210.2k & 157.2k & 230.0k & 492.7k \\
Bulgarian & 77.8M & 270.1k & 190.0k & 649.7k & 2922.6k \\
Catalan & 263.1M & 681.5k & 496.3k & 5529.4k & 7918.1k \\
Chinese & 408.3M & 1225.1k & 847.8k & 3339.8k & 4114.9k \\
Croatian & 53.0M & 183.2k & 110.2k & 590.3k & 849.3k \\
Czech & 168.1M & 472.3k & 363.0k & 2349.1k & 4182.0k \\
Danish & 68.2M & 271.8k & 212.3k & 2985.6k & 3313.5k \\
Dutch & 316.3M & 1984.7k & 910.8k & 18771.0k & 21404.2k \\
English & 2879.2M & 6043.6k & 4629.4k & 25146.2k & 28708.8k \\
Estonian & 41.2M & 217.2k & 147.3k & 1962.7k & 2254.0k \\
Finnish & 105.2M & 508.2k & 385.9k & 3249.4k & 5624.6k \\
French & 954.7M & 2271.5k & 1736.6k & 11125.7k & 13929.3k \\
Galician & 61.7M & 176.9k & 131.7k & 3423.2k & 5632.3k \\
German & 1022.7M & 2313.5k & 1888.3k & 13908.3k & 16704.8k \\
Hebrew & 165.7M & 303.6k & 213.0k & 3008.6k & 4901.6k \\
Hindi & 38.3M & 151.5k & 100.9k & 809.3k & 1039.4k \\
Hungarian & 142.1M & 475.8k & 366.7k & 2088.5k & 2391.2k \\
Indonesian & 113.2M & 600.3k & 345.3k & 2272.7k & 3681.5k \\
Irish & 7.0M & 56.3k & 37.3k & 4303.4k & 6605.3k \\
Italian & 598.9M & 1620.2k & 1328.9k & 6796.7k & 9388.0k \\
Japanese & 772.2M & 1229.6k & 921.2k & 2625.2k & 3271.9k \\
Korean & 94.9M & 534.4k & 362.4k & 507.0k & 875.7k \\
Latin & 11.2M & 132.3k & 98.5k & 311.9k & 2591.2k \\
Latvian & 23.1M & 106.2k & 79.9k & 463.0k & 692.3k \\
Lithuanian & 35.3M & 192.9k & 140.7k & 405.7k & 632.3k \\
Maltese & 3.3M & 4.9k & 3.6k & 77.2k & 225.6k \\
Marathi & 8.9M & 65.0k & 38.8k & 639.7k & 792.2k \\
Modern Greek (1453-) & 95.2M & 204.8k & 149.3k & 1552.8k & 1827.0k \\
Northern Sami & 0.3M & 7.9k & 4.6k & 148.6k & 242.1k \\
Norwegian Nynorsk & 28.5M & 158.7k & 122.1k & 2747.5k & 4106.0k \\
Persian & 123.8M & 900.1k & 704.4k & 1990.3k & 2512.9k \\
Polish & 305.0M & 1435.7k & 1138.5k & 2882.9k & 5337.8k \\
Portuguese & 346.4M & 1071.0k & 744.8k & 4829.5k & 7324.7k \\
Romanian & 92.1M & 407.6k & 301.4k & 3599.9k & 5903.1k \\
Russian & 647.0M & 1647.2k & 1312.5k & 3335.1k & 5936.4k \\
Scottish Gaelic & 1.7M & 15.6k & 10.0k & 274.7k & 435.4k \\
Serbian & 113.3M & 615.4k & 459.9k & 654.2k & 990.8k \\
Slovak & 41.0M & 209.7k & 144.5k & 1079.6k & 1366.4k \\
Slovenian & 49.8M & 161.8k & 117.6k & 3415.8k & 3642.6k \\
Spanish & 784.0M & 1685.9k & 1204.2k & 10575.4k & 13264.9k \\
Swedish & 229.0M & 2458.1k & 975.6k & 5361.9k & 7110.7k \\
Tamil & 27.7M & 147.2k & 96.2k & 794.8k & 1870.6k \\
Telugu & 33.2M & 71.6k & 52.9k & 731.3k & 854.1k \\
Turkish & 86.8M & 494.6k & 340.4k & 833.9k & 1184.3k \\
Uighur & 2.8M & 7.4k & 3.0k & 3.9k & 57.4k \\
Ukrainian & 311.2M & 1106.6k & 831.1k & 3879.9k & 6361.9k \\
Urdu & 30.6M & 175.0k & 137.0k & 746.1k & 917.3k \\
Vietnamese & 160.6M & 1256.2k & 365.9k & 660.3k & 2932.5k \\
Wolof & 0.6M & 1.6k & 0.9k & 222.8k & 292.2k \\
   \bottomrule
    \end{tabular}
    }
    \caption{Statistics of individual languages in DaMuEL. Tokens represent a number of tokens present in the language. Pages correspond to the number of Wikipedia pages. The number of Named entities with Page represents named entities that have Wikipedia pages. The number of named entities represents named entities that have at least some Wikidata info like label, description, or alias. Entities represent all items that have at least some Wikidata info like label, description, or alias.}
    \label{tab:languages}
\end{table*}

\begin{figure*}[t]
\begin{center}
\includegraphics[width=1.0\textwidth]{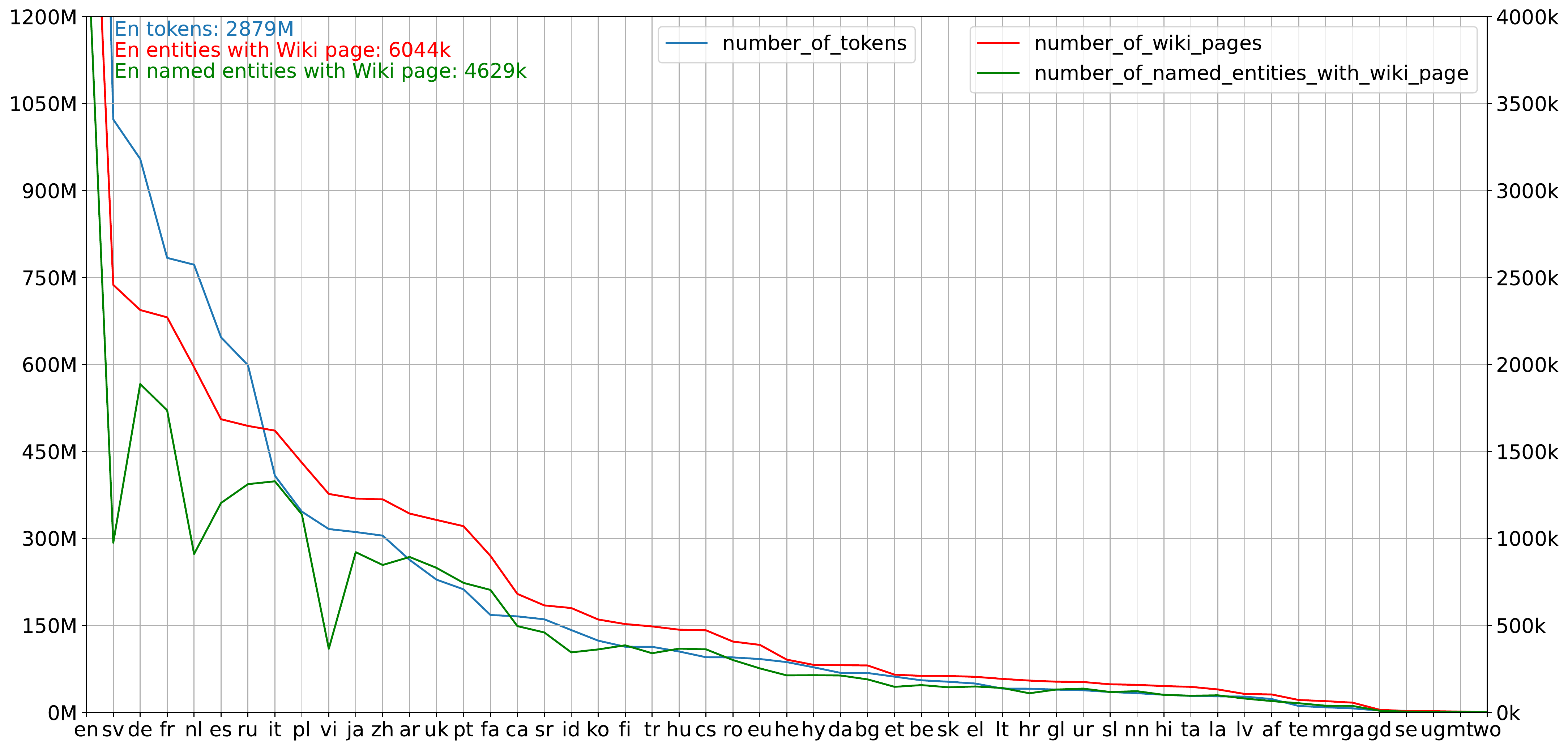}
\end{center}
\caption{The number of words and entities of the individual DaMuEL languages data.}
\label{fig.text_sums}

\bigskip
\bigskip

\begin{center}
\includegraphics[width=1.0\textwidth]{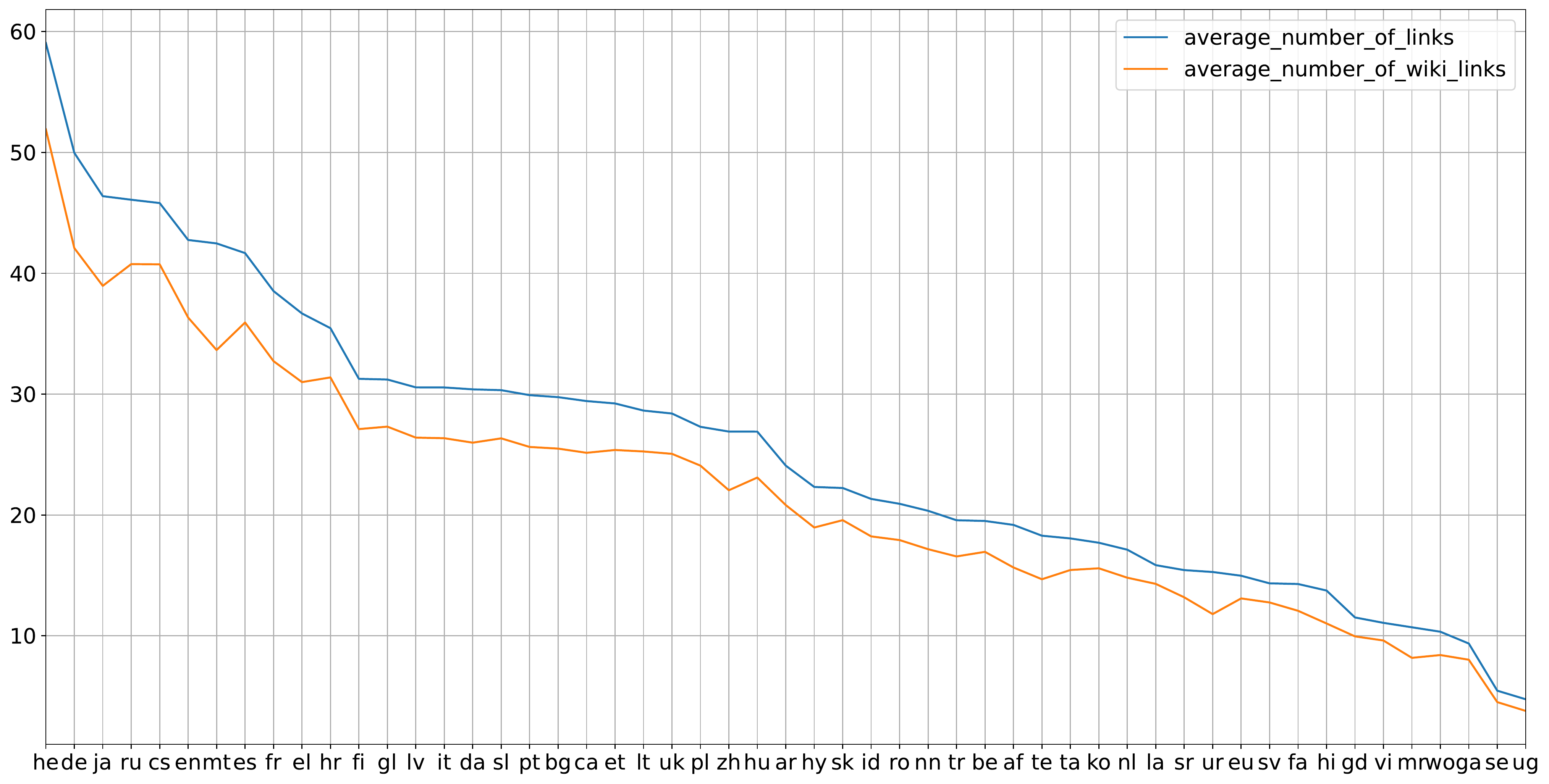}
\end{center}
\caption{The average number of original named entity links for every DaMuEL language, and the average number of flat expanded links.}
\label{fig.links_means}
\end{figure*}

\begin{figure*}[t]
\begin{center}
\includegraphics[width=1\hsize]{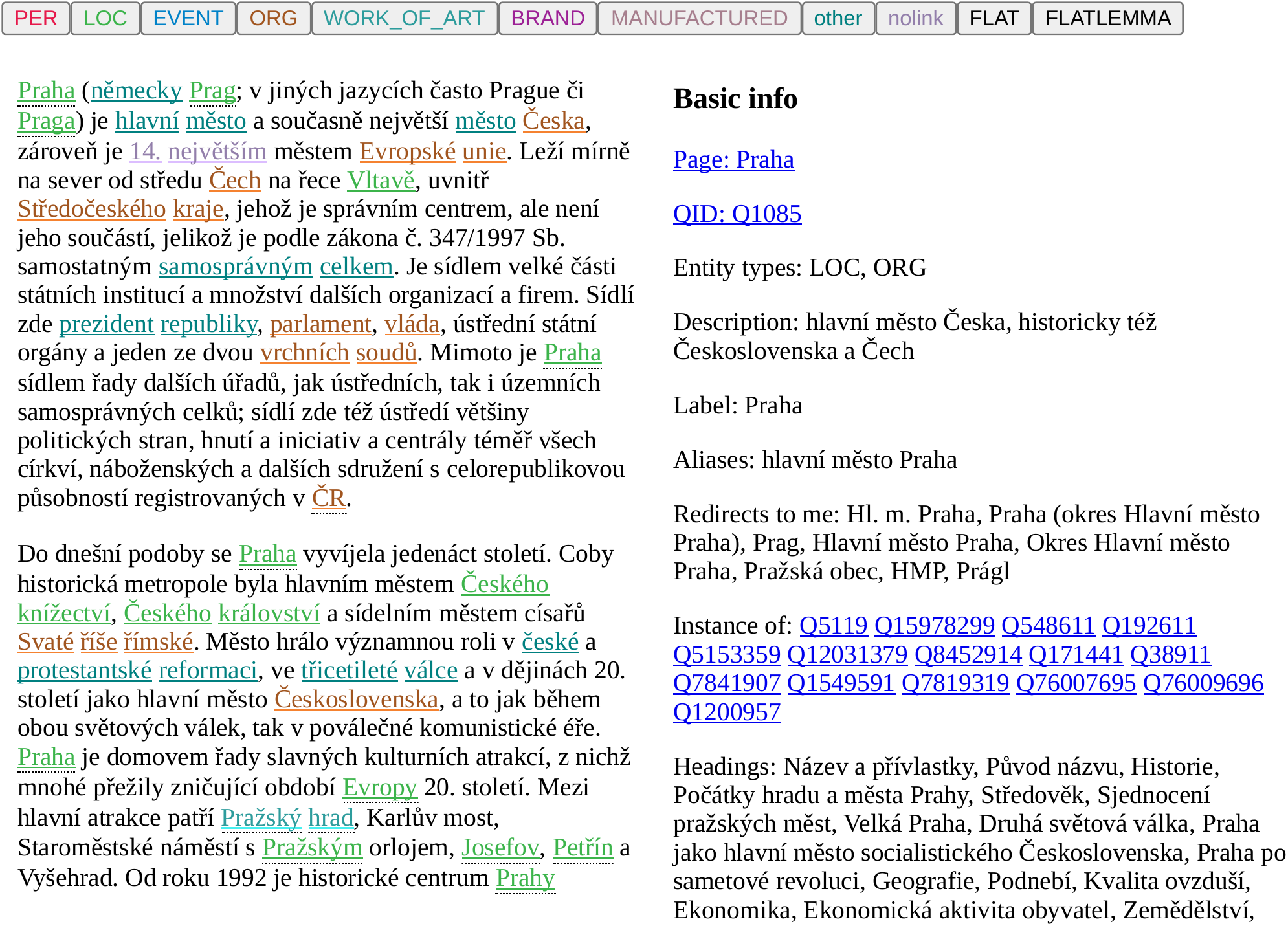} 
\caption{Screenshot of the DaMuEL visualization, showing the text with the flat expanded links, their named entity types, and all information about the entity itself.}

\label{fig.vis1}
\end{center}
\end{figure*}

We created DaMuEL using the Wikidata dump from Nov 9, 2022, and Wikipedia dumps that were the newest on Nov 1, 2022. The dataset is stored in a compressed JSON Lines format.\footnote{\scriptsize\url{https://jsonlines.org/}} Every entity is saved on a single line as a JSON dictionary.

We store the language-independent part of the knowledge base separately as DaMuEL Wikidata. It consists of a collection of entities containing a QID, all claims, statements, and ranks from Wikidata, and named entity types, if any. The schema describing the JSON structure of every such entity is presented in Figure~\ref{fig.wikidata_structure}.

All other information in DaMuEL is language specific, and we store the data for every language separately. Each entity includes a QID, localized names from Wikidata (labels, descriptions, and aliases), and information from the corresponding Wikipedia page, if available in the given language. For every Wikipedia page, we include the following:
\begin{citemize}
    \item it's title (which can also be used to construct the URL of the given Wikipedia page),
    \item it's text (which has been tokenized and morphologically analyzed),
    \item the original and expanded entity mentions (each consisting of a span, title, and a QID if it was possible to resolve it).
\end{citemize}
The structure is presented in Figure~\ref{fig.wiki_structure}.

The common identifier between language-independent and language-specific parts is the QID, which comes from Wikidata and links these two parts together.

Every language-specific dataset is stored in the tar archive, which consists of 500 parts, and each of these parts is compressed using xz. The language-independent part is stored in the same way.

The overall statistics of the dataset are quantified in Table~\ref{tab:statistics}. The whole dataset consists of 12.3G tokens, and its uncompressed size is 1.4TB on the disk. On average, there are nearly 500k entities that have Wiki page per language.

We also provide per-language statistics. The sizes of data for individual languages are presented in Table~\ref{tab:languages}
and in Figure~\ref{fig.text_sums}. Generally, the English Wikipedia is considerably larger than all others, but more than 20 languages contain data of at least 100M words and 500k wiki pages.

Finally, the average number of named entity links per page is displayed for every language in Figure~\ref{fig.links_means}.
There are, on average, 21.9 links to named entities per wiki page. With the expansion, there are 25.5 links. On average, across all languages, 17.2\% more links are created during the expansion.

\subsection{Visualization}
You can see our internal visualization of a Czech Wikipedia page Prague in Figure~\ref{fig.vis1}.

The page is divided into two columns and a button bar on the top, where the latter allows highlighting of certain entity types. In the left column, there is a Wikipedia text with links. In the right column, the Wikidata information is shown, plus additional information like all links on the page (allowing to highlight mentions of chosen entities), redirects to the entity, entity anchors, and others.

\subsection{Working with DaMuEL in Spark}
The distributed dataset compressed in the \texttt{xz} format can be loaded in Spark directly and transparently decompressed in a distributed manner. However, additional configuration is necessary. First, we need to add the following configuration when creating the \texttt{SparkContext}, which we illustrate in PySpark:
\begin{lstlisting}[language=python]
  import pyspark
  conf = pyspark.SparkConf()
  conf.set(
    "spark.hadoop.io.compression.codecs",
    "io.sensesecure.hadoop.xz.XZCodec")
  sc = pyspark.SparkContext(conf=conf)
\end{lstlisting}
Alternatively, the same effect can be achieved by defining the following environmental variable:

\begin{lstlisting}[basicstyle=\scriptsize\ttfamily]
SPARK_SUBMIT_OPTS="-Dspark.hadoop.io.compression.codecs=
io.sensesecure.hadoop.xz.XZCodec"
\end{lstlisting}

Finally, we need the XZ decompression package.
Spark can download is automatically by adding the following \verb|packages| parameter to \texttt{spark-submit}:
\begin{lstlisting}[basicstyle=\small\ttfamily]
  --packages "io.sensesecure:hadoop-xz:1.4"
\end{lstlisting}

\section{Conclusion and Future Work}
We presented DaMuEL, a massive multilingual dataset created from Wikidata and Wikipedia, which for 53 languages, provides a knowledge base and texts with annotated entity mentions. Even if our target application is a named entity linking system, the dataset contains a large quantity of tokenized and morphologically annotated Wikipedia texts, which could be useful on its own accord. The dataset is published under the CC BY-SA license at {\small\url{https://hdl.handle.net/11234/1-5047}}.

In the future, we naturally plan to experiment with the training and evaluation of named entity recognition and named entity linking systems, both for individual languages and with a multilingual model.

%

\section{Bibliographical References}\label{reference}

\bibliographystyle{lrec2022-bib}
\bibliography{lrec2022-damuel}


\end{document}